\documentclass[letterpaper, 10 pt, conference]{ieeeconf}  

\IEEEoverridecommandlockouts                              

\usepackage{amsmath} 
\usepackage{amssymb}  

\usepackage{graphicx} 
\usepackage{tabularx} 
\usepackage{balance} 
\usepackage[whole]{bxcjkjatype} 

\usepackage{algorithm}
\usepackage{algorithmicx}
\usepackage{amsfonts}
\usepackage{bm}
\usepackage{here}
\usepackage{comment}
\usepackage{multirow}
\usepackage{cite}
\usepackage{multibib}

\usepackage{booktabs} 
\usepackage{siunitx}
\usepackage[hyphens]{url} 
\usepackage[hidelinks]{hyperref} 
\usepackage[switch]{lineno} 

\usepackage{colortbl} 
\usepackage{xcolor}   

\definecolor{lightgray}{gray}{0.9} 

\usepackage{enumerate}
\usepackage{tcolorbox}

\usepackage[absolute,overlay]{textpos}
\title{\LARGE \bf
Take That for Me: Multimodal Exophora Resolution \\with Interactive Questioning for Ambiguous Out-of-View Instructions
}

\author{
    Akira Oyama$^{1}$,
    Shoichi Hasegawa$^{1,*}$,
    Akira Taniguchi$^{1}$,
    Yoshinobu Hagiwara$^{1, 2}$,
    and Tadahiro Taniguchi$^{1, 3}$
    \thanks{
        This work was supported by JSPS KAKENHI Grants-in-Aid for Scientific Research (Grant Numbers JP23K16975 and JP22K12212) and JST Moonshot Research \& Development Program (Grant Number JPMJMS2011).
    }
    \thanks{
        $^{1}$Akira Oyama, Shoichi Hasegawa, Akira Taniguchi, Yoshinobu Hagiwara, and Tadahiro Taniguchi are with Ritsumeikan University;
        2-150 Iwakura, Ibaraki, Osaka 567-8570, Japan.
        $^{2}$Yoshinobu Hagiwara is with Soka University;
         1-236 Tangi, Hachioji, Tokyo, 192-8577, Japan.
        $^{3}$Tadahiro Taniguchi is with Kyoto University;
        Yoshida Honmachi, Sakyo, Kyoto, Kyoto 606-8317, Japan.
        {\tt\small\{oyama.akira, hasegawa.shoichi, a.taniguchi\} @em.ci.ritsumei.ac.jp}, 
        {\tt\small hagiwara@soka.ac.jp},
        {\tt\small taniguchi@i.kyoto-u.ac.jp}
    }
    \thanks{
        $^{*}$Corresponding author.
    }
}

\begin{document}


\maketitle




\begin{abstract}
    %

    Daily life support robots must interpret ambiguous verbal instructions involving demonstratives such as ``Bring me that cup,'' even when objects or users are out of the robot's view. 
    Existing approaches to exophora resolution primarily rely on visual data and thus fail in real-world scenarios where the object or user is not visible. 
    We propose Multimodal Interactive Exophora resolution with user Localization (MIEL), which is a multimodal exophora resolution framework leveraging sound source localization (SSL), semantic mapping, visual-language models (VLMs), and interactive questioning with GPT-4o. 
    Our approach first constructs a semantic map of the environment and estimates candidate objects from a linguistic query with the user's skeletal data.
    SSL is utilized to orient the robot toward users who are initially outside its visual field, enabling accurate identification of user gestures and pointing directions. 
    When ambiguities remain, the robot proactively interacts with the user, employing GPT-4o to formulate clarifying questions. 
    Experiments in a real-world environment showed results that were approximately 1.3 times better when the user was visible to the robot and 2.0 times better when the user was not visible to the robot, compared to the methods without SSL and interactive questioning.
    The project website is \href{https://emergentsystemlabstudent.github.io/MIEL/}{\textcolor{blue}{https://emergentsystemlabstudent.github.io/MIEL/}}.
\end{abstract}


\begin{textblock*}{180mm}(15mm,\dimexpr\paperheight-15mm\relax)
{\footnotesize \copyright~2025 IEEE.  Personal use of this material is permitted. 
Permission from IEEE must be obtained for all other uses, in any current or future media,
including reprinting/republishing this material for advertising or promotional purposes,
creating new collective works, for resale or redistribution to servers or lists, 
or reuse of any copyrighted component of this work in other works.}
\end{textblock*}

\section{Introduction}
\label{sec:introduction}

    In our daily life, we frequently use verbal instructions that include demonstratives, such as ``Take that for me,'' but for robots, the target object is often unclear and the user or object is often not in the robot's view.
    One of the challenges in the field of robotics is enabling daily life support robots to understand and execute tasks based on such instructions and situations~\cite{taniguchi2019survey}.
    To achieve this, implementing exophora resolution~\cite{park2023visual,oyama2023exophora} is essential.
    Exophora resolution involves identifying the referent—whether a person or object—associated with anaphora (demonstratives or pronouns) within utterances, based on the surrounding context of the speaker or listener.
    For instance, if a user instructs the robot to ``Bring me that cup,'' the robot must identify the target object corresponding to ``that cup,'' even if there are many cups in the environment.

    When giving instructions that include demonstratives, people often use gestures such as pointing to indicate the direction in which the target object is located~\cite{lin2023gesture,nakagawa2024pointing}.
    Therefore, the robot must face the user's direction and consider gestures.
    However, there are many cases where the gesture cannot be observed because it is in the camera's blind spot, and the target of the instruction cannot be identified based on visual information alone.
    In such situations, the direction of speech is an important clue.
    

    Recent advancements in object detection and vision language models (VLMs)~\cite{radford2021learning,oquab2024dinov2} have been reported in numerous studies within the field of natural language processing (NLP), which incorporate not only textual data but also images and other modalities~\cite{yu2019you,yu2021exophoric,kang2019dual,yu2022vd,inadumi2024gaze}.
    Some studies investigated exophora resolution, a process that identifies images containing objects and individuals corresponding to target words in the text~\cite{yu2019you,yu2021exophoric}.
    These studies require the target object or person to be included in the image given as input.
    However, it is difficult to use these models as they are for a robot because the robot's view is limited, and there are many possible cases where the target object is placed outside the robot's view in the home environment.

    \begin{figure}[tb]
        \centering
        \includegraphics[width=\linewidth]{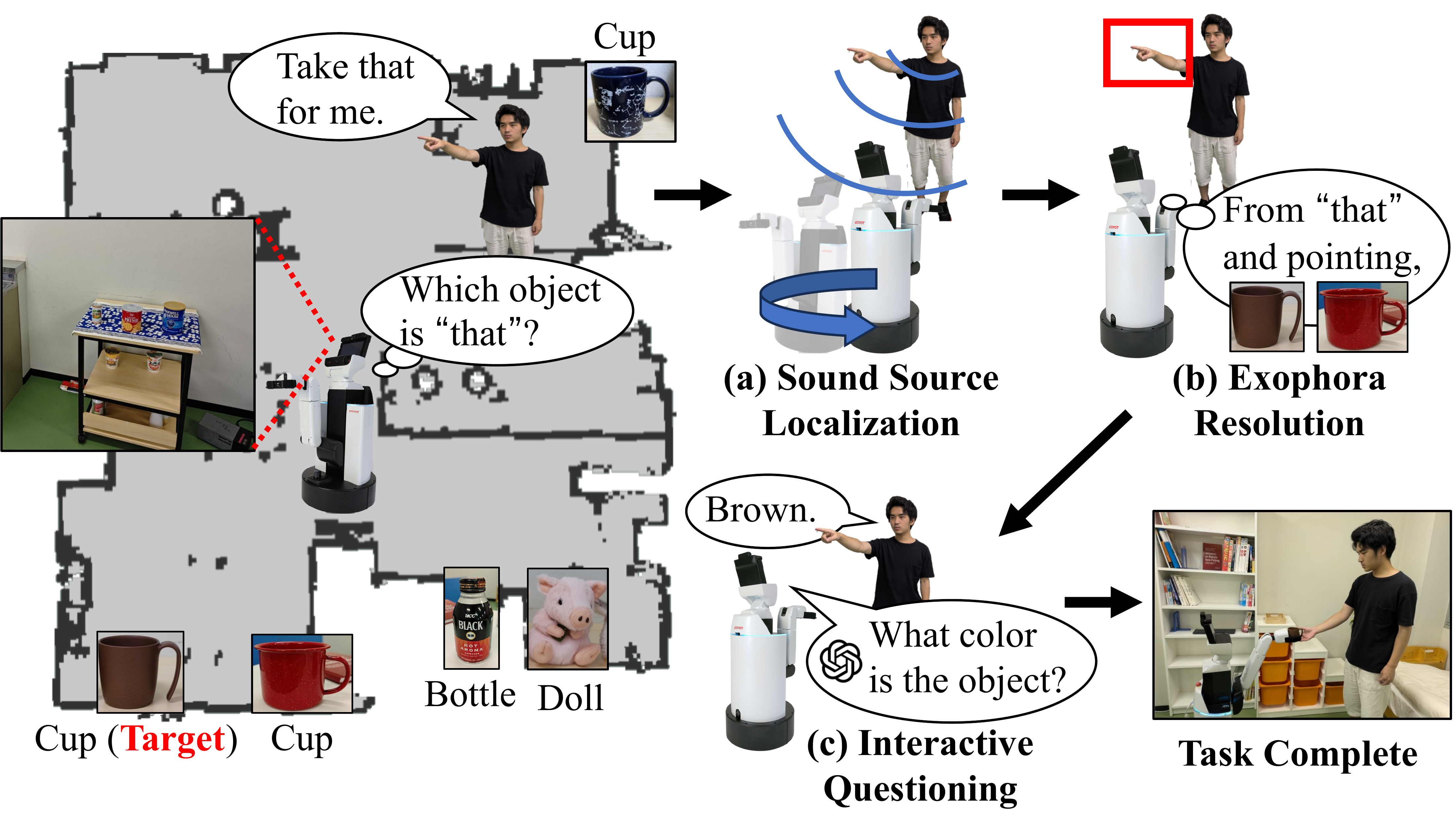}
        \caption{
        This study addresses situations where a user gives a robot an ambiguous command with a demonstrative (\textit{e.g.}, ``Take that for me''), especially when the user is outside the robot's view and non-verbal cues like pointing are unavailable, making it hard to identify what ``that'' refers to (Left).
        To address this issue, (a) Utilizing sound source localization to estimate the direction of the user and obtain pointing information. (b) Narrowing down candidate objects by exophora resolution. 
        (c) Supplementing missing information by asking questions using GPT-4o. 
        }
        \label{fig:abst}
    \end{figure}

    On the other hand, exophora resolution's research has been previously investigated in scenarios where the target object is out of the robot's field of view.
    An example is ECRAP, which enabled a robot to consider objects outside its field of view by leveraging pointing information, previously acquired object location data, and the characteristics of demonstratives~\cite{oyama2024ecrap}.
    This method incorporated the three types of Japanese demonstratives—{\it ko}-series, {\it so}-series, and {\it a}-series—and identifies the target object based on the spatial relationship between the speaker (user) or the listener (robot) and the object.
    While ECRAP enabled exophora resolution, it still has several limitations.
    One issue is that the success rate of object identification significantly drops when the linguistic query lacks class information about the target object or when the user is located outside the robot's field of view.
    Another limitation is that ECRAP cannot handle visual attributes such as color or shape mentioned in the linguistic query.

    Our study aims to achieve robust real-world exophora resolution that is resilient to incomplete observational data. 
    We show the task scenario and focused challenges of our study in Fig.~\ref{fig:abst}.
    In this study, the robot is given a linguistic query containing demonstratives such as ``Take that for me.''
    There are two major problems in performing exophora resolution in the real world:
    \begin{itemize}
        \item \textbf{Ambiguity and variability of user instructions}: Ambiguity due to demonstratives, lack of information about the object's appearance or class, and differences in expression by the speaker
        \item \textbf{User instructions from outside the robot's field of view}: Robots have difficulty relying on user location information to identify objects
    \end{itemize}
    These real-world problems of exophora resolution are discussed in detail in Section~\ref{sec:problem statement}.
    

    In this study, we propose \textbf{Multimodal Interactive Exophora resolution with user Localization (MIEL)}, which is a multimodal exophora resolution framework leveraging sound source localization (SSL), semantic mapping, VLMs, and interactive questioning.
    Firstly, the robot utilizes SSL to locate users outside its field of view (Fig.~\ref{fig:abst}(a)).
    Secondly, the robot performs an exophora resolution.
    The user's linguistic query is converted into feature representations using GPT-4o~\cite{hurst2024gpt} as large language models (LLMs) and VLMs.
    By combining the object-related information obtained from the linguistic query with the user's skeletal data, the system estimates a ``demonstrative region'' where the target object is likely to exist and predicts the object within that region.
    Furthermore, the direction indicated by the user's pointing gesture is used to identify the target object.
    When the user is outside the robot's field of view, the system utilizes SSL to estimate the user's direction and adjusts the robot's orientation accordingly to acquire skeletal data.
    VLMs and semantic maps are also employed to interpret visual attributes such as color and shape, mentioned in the linguistic query (Fig.~\ref{fig:abst}(b)).
    Finally, if the robot is unable to narrow down the object after completing the exophora resolution, it performs interactive questioning to supplement the information (Fig.~\ref{fig:abst}(c)).

    Our contribution is three-fold:
    \begin{enumerate}
        \item We demonstrated the effectiveness of SSL for acquiring user skeletal data and pointing gestures in scenarios where users are initially outside the robot's field of view, revealing its crucial role in multimodal exophora resolution.
        \item We clarified the importance of interactive questioning for resolving ambiguous referring expressions lacking explicit object class or feature in linguistic instructions.
        \item We identified conditions under which semantic mapping and VLMs effectively resolve referential ambiguities in linguistic queries, highlighting their complementary role within multimodal exophora resolution.
    \end{enumerate}

    This paper is structured as follows:
    Section~\ref{sec:problem statement} describes the problem statement.
    Section~\ref{sec:related works} describes related works.
    Section~\ref{sec:proposed method} presents our proposed method in detail.
    Section~\ref{sec:experiment} describes the experiments.
    Lastly, Section~\ref{sec:conclusion} summarizes this study and discusses future works.

\section{Problem Statement}
\label{sec:problem statement}
    In this study, we address the task of real-world exophora resolution for daily life support robots.
    Real-world scenarios present several critical and challenging issues that robots must overcome:

\subsection{Ambiguity and Variability of User Instructions}
\label{subsec:various instraction}
    Daily life support robots frequently encounter linguistic instructions that are inherently ambiguous and incomplete, such as ``Bring me that cup,'' or ``Take this.''
    Such instructions often lack explicit information about object classes, features, or precise identification cues.
    In other words, referring expressions are ambiguous, including demonstratives and the lack of information about the appearance and class of objects. 
    Furthermore, combining these ambiguous elements and the variability of expressions across different speakers lead to diverse and inconsistent ways of conveying the same intent.
    To function effectively in real-world environments, robots must robustly infer the user's intent and correctly identify target objects despite limited and vague linguistic inputs.

    When instructions are ambiguous, the ambiguity can be resolved by asking questions.
    For instance, vague instructions like ``Bring me that'' lack critical class or feature details, compelling the robot to proactively ask clarifying questions. 
    Hence, the ability to effectively interact with users and actively obtain additional information is essential for achieving robust and accurate exophora resolution in real-world environments.

\subsection{User Instructions from Outside the Robot's Field of View}
\label{subsec:problem statement:out-of-view}
    In realistic home environments, users often give instructions from positions outside the robot's immediate field of view, posing significant challenges.
    Under these circumstances, robots cannot rely on the user's position information.
    
    Instead, robots must utilize non-visual modalities, including SSL and previously constructed semantic maps, to infer the user's location and estimate the positions of target objects.
    Effective exophora resolution thus requires advanced multimodal reasoning capabilities to handle these inherently uncertain and partially observable scenarios.



\section{Related Works}
\label{sec:related works}
This section reviews existing research on exophora resolution in NLP and robotics, highlighting their limitations. 

\subsection{Exophora Resolution in NLP}
    Most of the research on exophora resolution in the field of NLP is based on identifying target objects in an image from the image and text~\cite{yu2019you,yu2021exophoric,kang2019dual,yu2022vd,ueda2024jcre3,inadumi2024gaze}.
    Pronoun coreference resolution (PCR) was introduced to identify nouns referred to by pronouns within dialogue scenes, accompanied by constructing the dialogue PCR dataset called VisPro~\cite{yu2019you}.
    Additionally, VisCoref tackles the PCR task using the VisPro dataset.
    However, it is difficult to use the models of exophora resolution in the field of NLP for robots because the target object is not always located in the image obtained from a robot with a limited field of view.

    Recently, a study in NLP has integrated multimodal information, such as images and gaze data~\cite{inadumi2024gaze}. 
    In a related effort, the GazeVQA dataset was developed, focusing not only on directives and pronouns but also considering gaze information from images~\cite{inadumi2024gaze}.
    This approach further leverages gaze data to enhance visual question answering (VQA) performance, demonstrating the utility of multimodal information in resolving ambiguous references.
    In this study, we attempt to actively acquire information outside the robot's field of view from sound sources and pointing cues.

    
    
\subsection{Exophora Resolution in Robotics}
    Some research addressing exophora resolution assume a tabletop environment, where the target object must be present within the image~\cite{lin2023gesture,wang2024language}.
    GIRAF processes images of human gestures and linguistic queries to plan robot actions~\cite{lin2023gesture}.
    This method comprises three modules: scene descriptor, human descriptor, and LLM task planner. 
    It predicts the target object based on queries such as ``Give me that tool.''
    On the other hand, exophora resolution frameworks for a mobile-based daily life support robot operating in a real environment~\cite{oyama2023exophora} have been proposed.
    The robot performs exophora resolution using inputs from multiple sources, including object location and classification obtained through prior environment exploration, linguistic instructions from the user, and images of the user taken from the robot's view.
    ECRAP, an application of this framework, combines a task classification module and LLM to do the planning expected by users based on their various linguistic queries~\cite{oyama2024ecrap}.

    The previously proposed framework for exophora resolution can consider objects beyond the robot's view that are not in the image~\cite{oyama2023exophora,oyama2024ecrap}. 
    However, this approach has two problems.
    Firstly, if the user's skeletal information is missing, the success rate is greatly reduced. 
    To address this, our method incorporates SSL as a complementary cue.
    Secondly, the problem is the inability to consider words that describe features such as the color and shape of objects in the linguistic query.
    To address this, we utilize semantic maps and VLMs to improve the understanding of the referring expressions in the linguistic query.

\subsection{Interactive Questioning}
\label{subsec:interactive questioning}
    Ambiguities in the user's instructions can be resolved by enabling robots to ask clarifying questions to the user~\cite{knowno2023,park2023clara}.
    KNOWNO is a method for planning robot actions using LLMs~\cite{knowno2023}.
    This method resolves ambiguities during planning by asking the user clarifying questions, thus determining the robot's subsequent actions.
    CLARA is an uncertainty estimation method~\cite{park2023clara} that uses LLM to classify commands given to robots as either ``clear instructions'' or ``ambiguous or impossible instructions.''
    If CLARA determines that a command is ambiguous, it uses LLM to ask the user additional questions and resolves the ambiguity of the command through dialogue.

    Our research uses to a method in which CLARA uses LLM to classify ambiguous instructions and resolves ambiguity by generating questions.
    Specifically, we first infer whether the target object is identified through exophora resolution.
    If identification fails, we use LLM to generate clarifying questions to resolve the ambiguity.



    \begin{figure*}[tb]
        \centering
        \includegraphics[width=\linewidth]{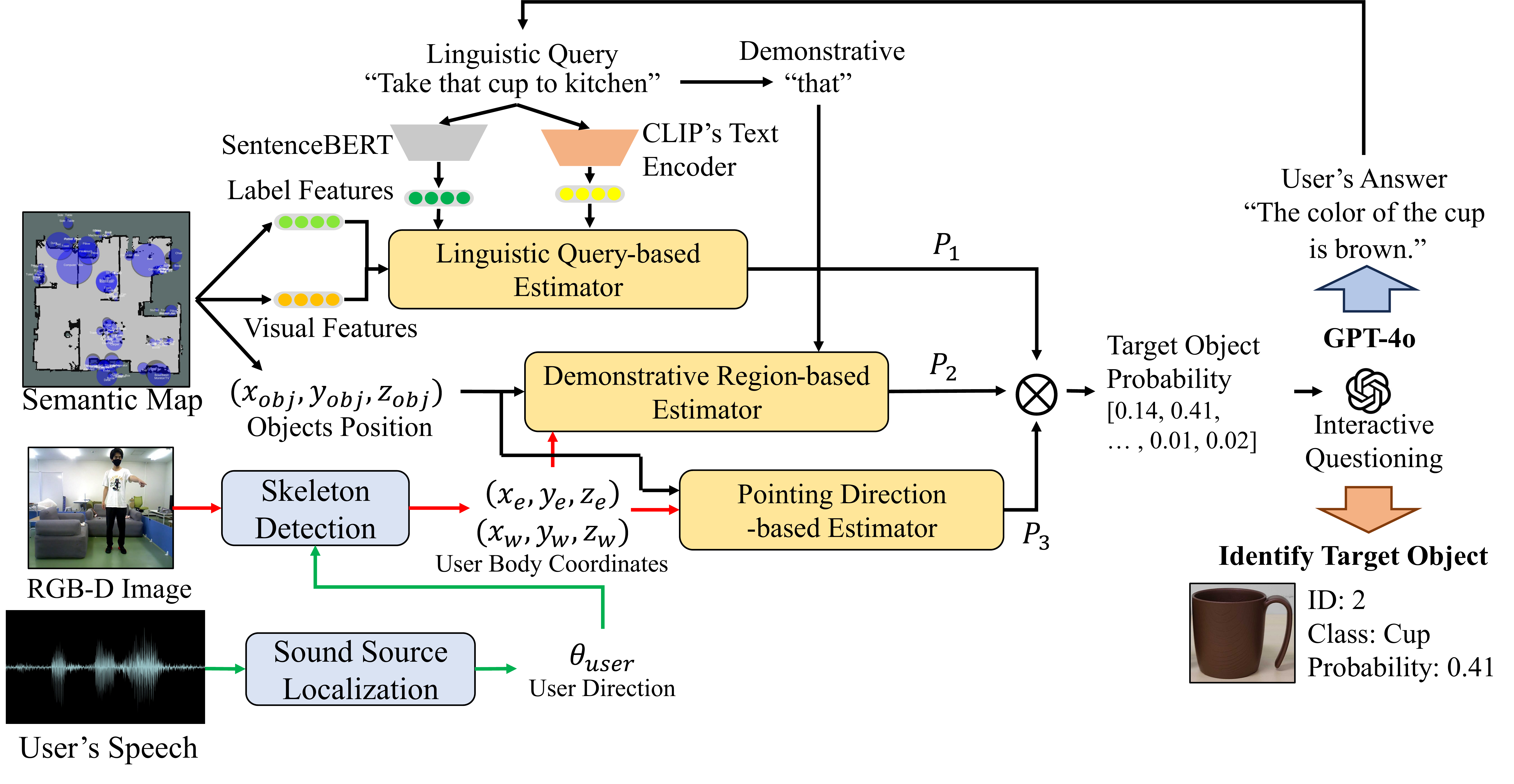}
        \caption{
        An overview of the MIEL. Initially, linguistic queries from the user, including demonstratives, are processed into semantic and visual features using SentenceBERT and CLIP encoders. A semantic map provides object locations and visual data, while skeletal detection and SSL determine user direction and pointing gestures. Three estimators generate candidate target object probabilities. If the initial identification is ambiguous, GPT-4o engages the user interactively through targeted questions, refining object identification and ensuring robust exophora resolution.
        }
        \label{fig:proposed_overview}
    \end{figure*}

\section{Proposed Methods}
\label{sec:proposed method}
    The proposed method, which is called MIEL, realizes exophora resolution using a semantic map, images from the robot's field of view, and spoken utterances as input.
    We show an overview of the MIEL in Fig.~\ref{fig:proposed_overview}.


\subsection{Preprocessing for Estimators}
\label{subsec:proposed_preprocessing}
    The linguistic query is processed through SentenceBERT's encoder~\cite{reimers2019sentence} and CLIP text encoder~\cite{radford2021learning}, which extracts and outputs features.
    If the linguistic query is not in English, it is first translated into English using GPT-4o before being fed into the encoders. 
    If the query is already in English, it is used without modification.
    Additionally, GPT-4o extracts only the demonstratives from the linguistic query.

    The semantic map extends the NLMap~\cite{chen2023open} into 3D space.
    From NLMap, the label features, visual features, and 3D map coordinates of each object ($x_{obj}$, $y_{obj}$, $z_{obj}$) are acquired. 
    The label features correspond to object detector labels, which are obtained through the SentenceBERT encoder. 
    The visual features are derived by extracting bounding boxes detected by the object detector and processing them through CLIP's image encoder.

    The 3D map coordinates of the user's eyes ($x_e$, $y_e$, $z_e$) and wrists ($x_w$, $y_w$, $z_w$) are extracted from the RGB-D image captured from the robot's viewpoint using a skeletal detector. 
    The MediaPipe~\cite{lugaresi2019mediapipe} is used as a skeletal detector.
    A vector originating from the user's eyes and passing through the wrist is used to determine the user's pointing direction. 
    The user's speech is captured by a microphone array, and the user's direction is estimated through SSL.
    If the user is not visible in the RGB-D image from the robot's viewpoint, the robot reorients itself toward the user based on the direction estimated by SSL to acquire the user's skeletal information.

    This information is processed by three estimators: linguistic query-based estimator, demonstrative region-based estimator, and pointing direction-based estimator. 
    Each estimator outputs the probability that a given object in the environment is the target object. 
    The top five objects, determined by multiplying the probabilities from each estimator, are then input to GPT-4o.
    If GPT-4o successfully identifies the target object, the exophora resolution process terminates. 
    If identification fails, GPT-4o asks the user questions to gather additional hints about the target object and then reattempts the exophora resolution.
    The interactive questioning is limited to a single exchange. 
    If the target object is still not definitively identified in the second attempt, the object with the highest probability is selected as the final target.

    \begin{figure}[tb]
        \centering
        \includegraphics[width=\linewidth]{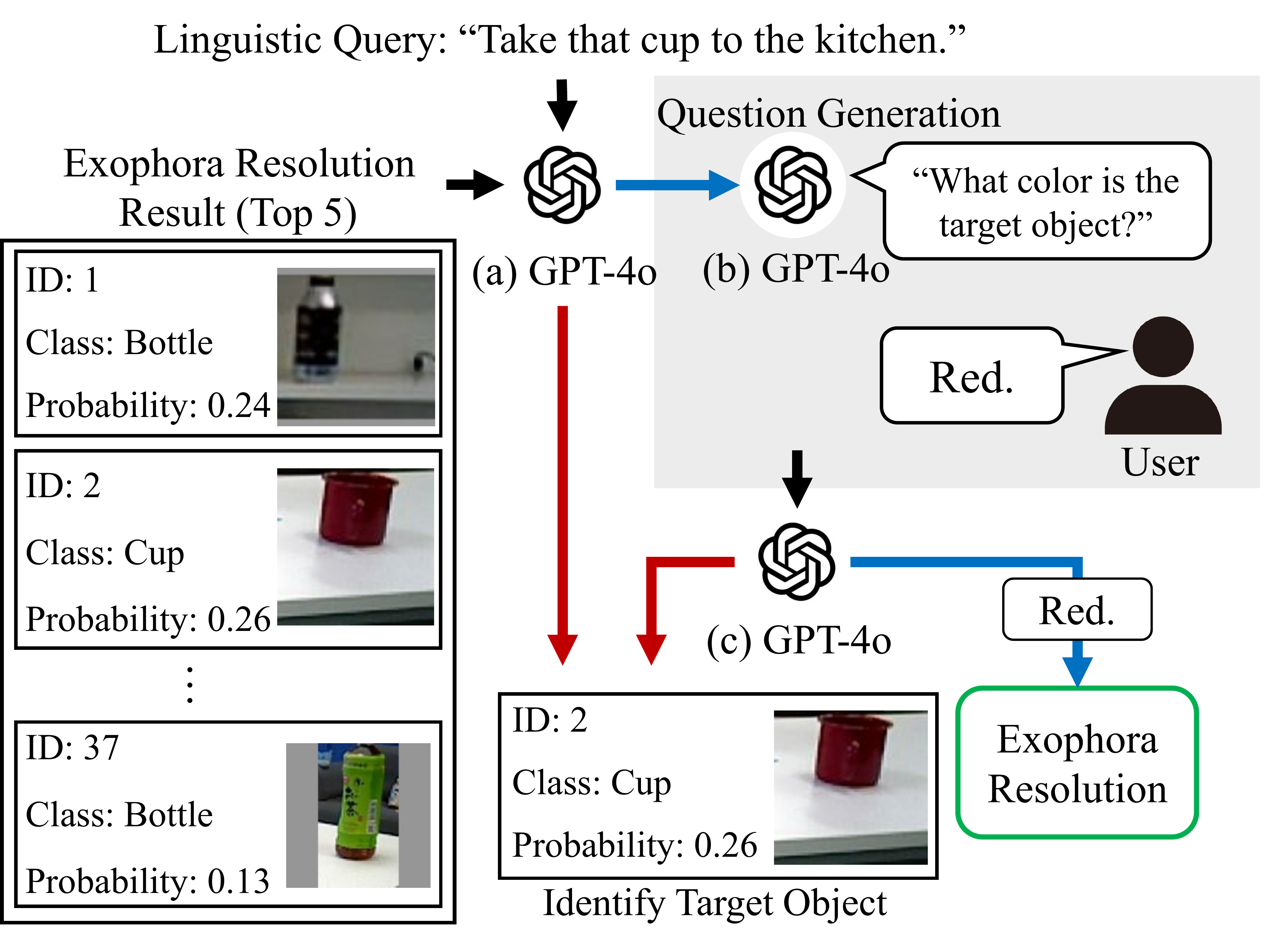}
        \caption{
        An overview of the module for interactive questioning.
        The red arrows represent the process flow when the target object is successfully identified, while the blue arrows indicate the process flow when the target object remains unidentified.
        }
        \label{fig:QandA_GPT4o}
    \end{figure}

\subsection{Linguistic Query-based Estimator}
\label{subsec:proposed_linguistic_query_based_estimator}
    The linguistic query-based estimator takes as input the object label features extracted from the semantic map, visual features, and linguistic query features obtained from the SentenceBERT and CLIP's text encoders. 
    It then calculates the probability $P_1$ that each candidate object in the environment is the target object.
    The estimator computes the cosine similarity between the label features of candidate objects extracted from the semantic map and the features obtained by processing the linguistic query through SentenceBERT's encoder.
    Similarly, the cosine similarity is computed between the features obtained by processing the visual features of candidate objects (extracted from the semantic map) and the linguistic query through CLIP's text encoder.
    The two cosine similarities are then multiplied and normalized to produce the final probability of each object being the target object.


\subsection{Demonstrative Region-based Estimator}
\label{subsec:proposed_demonstrative_based_estimator}
    The demonstrative region-based estimator takes the coordinates of objects extracted from the semantic map, the user's skeletal information obtained from the skeletal detector, and the demonstratives in the linguistic query as input. 
    It then calculates the probability $P_2$ that each candidate object in the environment is the target object.
    Utilizing the characteristics of the {\it ko}-, {\it so}-, and {\it a}-series of demonstratives, the demonstrative region for each series is modelled as a 3D Gaussian distribution.
    From the RGB-D image captured from the robot's viewpoint, the user's skeleton is detected using MediaPipe, and the mapped coordinates of the user's eyes and wrists are obtained.
    GPT-4o extracts a demonstrative from the linguistic query.
    This estimator specifically processes {\it ko}-, {\it so}-, and {\it a}-series demonstratives, excluding those from the {\it do}-series.
    For each {\it ko}-, {\it so}-, and {\it a}-series, the mean and variance of a three-dimensional Gaussian distribution representing the demonstrative regions are calculated. 
    The probability that each object is the target object is then determined based on the probability density function of the distribution.
    The formulas are the same as in this paper~\cite{oyama2023exophora}.

    The {\it ko}-series demonstratives refer to objects or persons close to the speaker; the mean of the 3D Gaussian distribution is set to the coordinates of the user's wrist.
    Similarly, {\it so}-series demonstratives indicate objects or persons near the listener, meaning the mean of the 3D Gaussian distribution is set to the coordinates of the robot.
    In contrast, {\it a}-series demonstratives are used for objects or persons far from both the speaker and the listener. 
    Thus, the mean of the 3D Gaussian distribution is set to the tip of the pointer.


\subsection{Pointing Direction-based Estimator}
\label{subsec:proposed_pointing_based_estimator}
    The pointing direction-based estimator inputs the coordinates of objects from the semantic map and the user's skeletal information from the skeletal detector. 
    It then calculates the probability $P_3$ that each candidate object in the environment is the target object based on the user's pointing direction.
    This estimator utilizes two types of vectors:
    Firstly, a vector originating from the user's eye and terminating at the user's wrist is referred to as the pointing vector.
    Secondly, a vector originating from the user's eye and terminating at each candidate object is referred to as the object direction vector.
    Since the object direction vector terminates at each candidate object, the number of such vectors corresponds to the number of objects.
    For each object, the angle $\theta_{n}$ between the pointing vector and the corresponding object direction vector is computed.
    This angle is then used to obtain the probability density from a two-dimensional von Mises distribution, where angles serve as random variables.
    The formulas are the same as in this paper~\cite{oyama2023exophora}.

\subsection{Prediction of Target Object with interactive questioning}
\label{subsec:proposed_QandA}
    After obtaining the probability of each object in the environment being the target object from the three estimators, we use this probability along with the GPT-4o~\footnote{See the project page (https://emergentsystemlabstudent.github.io/MIEL/) about the prompts in Fig.~\ref{fig:QandA_GPT4o} (a)--(c).} to predict the target object. 
    We show an overview of the module for interactive questioning in Fig.~\ref{fig:QandA_GPT4o}.
    Initially, GPT-4o receives the following information: the IDs of the top five objects with the highest probability of being the target object, their respective classes, the probability of each object being the target, corresponding images, and the user-provided query.
    If GPT-4o can determine the target object based on this information, it outputs only the ID of the identified object and terminates the process.
    On the other hand, if GPT-4o cannot identify the target object, it generates a clarifying question to facilitate identification.
    The user's responses to these generated questions are incorporated into the existing information, and the system attempts to predict the target object again.
    When the target object is successfully identified after integrating the user's answers, the system outputs only the object's ID and terminates the process.
    If the target object still cannot be identified, the user's responses are appended to the query, and exophora resolution is performed again.
    This interaction between the user and the system occurs only once. 
    The system does not reprocess the results of the second exophora resolution, as excessive interactions are time-consuming and impose an additional burden on the user.


\section{Experiments}
\label{sec:experiment}
In this experiment, we conducted the validation in a simulated home environment under a combination of whether or not the user was in the robot's field of view and the level of ambiguity of the instructions.

\subsection{Experimental Purpose}
\label{subsec:exp_abs}
    The purpose of the experiment is three-fold:
    \begin{enumerate}
        \item To verify how accurately exophora resolution identifies the correct target object.
        \item To verify how the results of exophora resolution vary depending on the amount of information provided in the linguistic query.
        \item To verify the contributions of SSL and interactive questioning to exophora resolution.
    \end{enumerate}
    To achieve the three objectives, we conducted experiments by preparing linguistic queries with different amounts of information, baselines, and a topline.

\subsection{Experimental Condition}
\label{subsec:exp_condition}
\begin{figure}[tb]
        \centering
        \includegraphics[width=\linewidth]{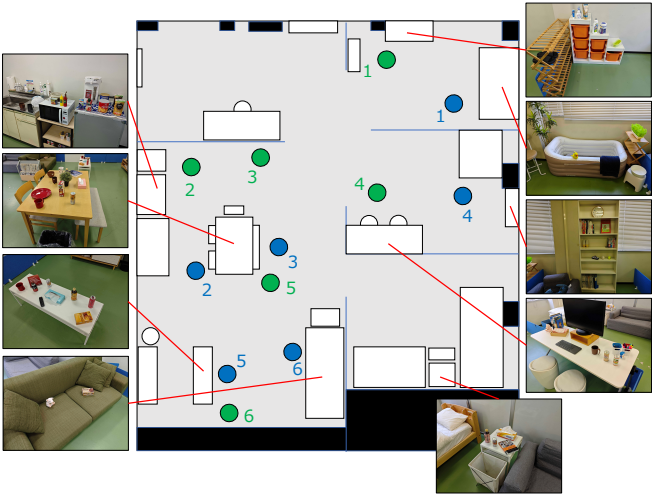}
        \caption{
        The layout of the experimental environment.
        The green circles are the user's location, and the blue circles are the robot's location. 
        }
        \label{fig:exp_user_robot_position}
    \end{figure}

    The experiment was conducted in a real-world setting (Fig.~\ref{fig:exp_user_robot_position}) designed to simulate a home environment.
    In Fig.~\ref{fig:exp_user_robot_position}, the green circle indicates the user's position, while the blue circle indicates the robot's position.
    When the numbers in the circles matched, the user pointed the robot toward the target object.
    Two types of the robot's initial positions were considered: one where the user was visible from the robot's perspective and another where the user was not.
    We compared the results between these two conditions when the user was visible and when the user was not visible from the robot's initial position. 

    The PC was equipped with an Intel Core i9-12900KF CPU, 64GB RAM, and a 10GB NVIDIA GeForce RTX 3080 for this experiment.
    In addition, this PC was built by the software environment for the robot~\cite{el2022software}, with Ubuntu 20.04 LTS and ROS Noetic version.
    ReSpeaker Mic Array v2.0 was used for the microphone array.
    The GPT-4o version used was \texttt{gpt-4o-2024-08-06}.

\subsubsection{Data Collection}
\label{subsubsec:data}
    The data was collected using the Human Support Robot (HSR)~\cite{yamamoto2019development} developed by Toyota Motor Corporation.
    The environment (Fig.~\ref{fig:exp_user_robot_position}) has several objects such as cups, books and plastic bottles, and the robot created an occupancy grid map and a semantic map by exploring the environment in advance.
    The semantic map stores information not only on small objects but also on larger items, such as beds and chairs.
    These objects were detected using Detic~\cite{zhou2022detecting}, with labels derived from the Objects365~\cite{shao2019objects365} dataset.
    A total of 114 objects across 39 classes were detected from 775 images during the creation of semantic maps in the environment.


    We collected the user's skeletal data and the results of SSL obtained based on spoken utterances in advance, and experiments were conducted using these data.
    In this experiment, we assumed that in MIEL, if SSL succeeds when the user is not visible from the robot's initial position, the robot can detect the skeleton by turning around.
    Conversely, if the SSL failed under the same conditions, the user's skeletal data could not be utilized.
    SSL was considered successful if the angular error was within 29 degrees of the correct direction and unsuccessful if the error exceeded 29 degrees.
    The 29-degree threshold corresponds to half of the horizontal field of view of the HSR camera, Xtion PRO LIVE, which is 58 degrees.
    This setting was made because the user can be seen within the robot's view within 29 degrees.
    
\subsubsection{Linguistic Queries}
\label{subsubsec:language_query}
    Linguistic queries were divided into the following three levels.
    \begin{itemize}
        \item Level~1: object class， demonstratives， query containing object features (\textit{e.g.}, ``Bring me that stuffed pig.'')
        \item Level~2: object class，query containing a demonstrative (\textit{e.g.}, ``Bring me that doll.'')
        \item Level~3: query containing only a demonstrative (\textit{e.g.}, ``Bring me that.'')
    \end{itemize}
    Object features represent attributes expressed through referential terms such as color, shape, and size.
    Level~1 queries were created first, followed by level~2 and level~3 queries, which were generated sequentially by removing information from the level~1 queries.

    The linguistic queries were created in Japanese, and the experiment was conducted.
    A total of 30 level~1 queries were created, five for each of the six user locations shown in Fig.~\ref{fig:exp_user_robot_position}.
    Level~2 and 3 queries, in which information was removed from level~1, were also tested with 30 queries each, for a total of 90 linguistic queries.

\subsection{Comparison Methods}
\label{subsec:exp_evaluation}

    We prepared the following comparison methods: VGPN~\cite{hu2018vgpn}, ECRAP~\cite{oyama2024ecrap}, Human (topline) and Human (w/o Q\&A).

    Firstly, VGPN is a method for identifying a target object based on two pieces of information: pointing and object class.
    VGPN is a method in which the object closest to the user's pointing and the object class is the same as the linguistic query is selected as the target object.
    Since VGPN does not handle a demonstrative, it cannot account for the positional relationship between the user and the robot.
    VGPN was used as a baseline for comparison to demonstrate the effectiveness of MIEL in handling demonstratives.

    Secondly, we use ECRAP.
    This method performs exophora resolution using an exophora resolution framework~\cite{oyama2023exophora} and LLMs.
    Although ECRAP is designed to perform action planning after identifying the target object, action planning was not conducted in this experiment, as the process concluded with object identification.
    ECRAP is capable of handling a demonstrative, but does not include SSL or a module for user interaction.
    In this experiment, ECRAP was used as a baseline for comparison to demonstrate the effectiveness of MIEL in handling SSL and interactive questioning.

    Thirdly, as a topline, human prediction of the target object is performed and compared with other methods (Human (topline)).
    In this experiment, a person stood at the robot's position (as shown in Fig.~\ref{fig:exp_user_robot_position}) and received instructions from another person positioned at the user's location to perform the exophora resolution.
    If the target object is unknown, it is also possible to ask clarifying questions to the person who gave the instructions.

    Finally, Human (w/o Q\&A) is used.
    The difference with Human (topline) is that it is not possible to ask questions to the speaker.
    In the experiment, the Human (w/o Q\&A) result was obtained by having the subject predict the target object without asking the speaker any questions.
    Subsequently, the Human (topline) result was obtained by allowing the subject to ask a question, receive an answer, and then predict the target object.
    By comparing the MIEL with Human (topline) and Human (w/o Q\&A), we evaluate the extent to which its success rate differs from that of humans in exophora resolution.

\begin{table*}[bt]
\centering
\caption{
The SR of exophora resolution when the user is visible from the robot's initial position.
}
\label{tbl:result_each_method_combined_seen}
\begin{tabular}{lcccccccc}
    \toprule
         & \multicolumn{4}{c}{\textbf{Top 1}} & \multicolumn{4}{c}{\textbf{Top 5}} \\ 
             \cmidrule(lr){2-5}
             \cmidrule(lr){6-9}
\textbf{Method}   & \multicolumn{1}{c}{\textbf{Level 1}} & \multicolumn{1}{c}{\textbf{Level 2}} & \multicolumn{1}{c}{\textbf{Level 3}} & \textbf{Total} 
         & \multicolumn{1}{c}{\textbf{Level 1}} & \multicolumn{1}{c}{\textbf{Level 2}} & \multicolumn{1}{c}{\textbf{Level 3}} & \textbf{Total} \\ 
             \midrule
VGPN~\cite{hu2018vgpn}    & \multicolumn{1}{c}{0.40 (12/30)} & \multicolumn{1}{c}{0.33 (10/30)} & \multicolumn{1}{c}{0.13 (\, 4/30)} & 0.28 (25/90)
         & \multicolumn{1}{c}{- (-)} & \multicolumn{1}{c}{- (-)} & \multicolumn{1}{c}{- (-)} & - (-) \\ 
ECRAP~\cite{oyama2024ecrap}    & \multicolumn{1}{c}{0.57 (17/30)} & \multicolumn{1}{c}{0.53 (16/30)} & \multicolumn{1}{c}{0.17 (\, 5/30)} & 0.42 (38/90)
         & \multicolumn{1}{c}{0.73 (22/30)} & \multicolumn{1}{c}{0.70 (21/30)} & \multicolumn{1}{c}{0.57 (17/30)} & 0.67 (60/90) \\ 
MIEL (ours) & \multicolumn{1}{c}{\textbf{0.63 (19/30)}} & \multicolumn{1}{c}{\textbf{0.60 (18/30)}} & \multicolumn{1}{c}{\textbf{0.37 (11/30)}} & \textbf{0.53 (48/90)}
                & \multicolumn{1}{c}{\textbf{0.87 (26/30)}} & \multicolumn{1}{c}{\textbf{0.87 (26/30)}} & \multicolumn{1}{c}{\textbf{0.63 (19/30)}} & \textbf{0.79 (71/90)} \\ \hline
Human (topline)    & \multicolumn{1}{c}{1.00 (30/30)} & \multicolumn{1}{c}{1.00 (30/30)} & \multicolumn{1}{c}{0.93 (28/30)} & 0.98 (88/90)
         & \multicolumn{1}{c}{1.00 (30/30)} & \multicolumn{1}{c}{1.00 (30/30)} & \multicolumn{1}{c}{1.00 (30/30)} & 1.00 (90/90) \\ 
Human (w/o Q\&A) & \multicolumn{1}{c}{0.97 (29/30)} & \multicolumn{1}{c}{0.93 (28/30)} & \multicolumn{1}{c}{0.67 (20/30)} & 0.86 (77/90)
                 & \multicolumn{1}{c}{1.00 (30/30)} & \multicolumn{1}{c}{1.00 (30/30)} & \multicolumn{1}{c}{0.97 (29/30)} & 0.99 (89/90) \\ 
    \bottomrule
\end{tabular}
\end{table*}

\begin{table*}[bt]
\centering
\caption{
The SR of exophora resolution when the user is not visible from the robot's initial position.
}
\label{tbl:result_each_method_combined_unseen}
\begin{tabular}{lcccccccc}
    \toprule
         & \multicolumn{4}{c}{\textbf{Top 1}} & \multicolumn{4}{c}{\textbf{Top 5}} \\ 
             \cmidrule(lr){2-5}
             \cmidrule(lr){6-9}
\textbf{Method}   & \multicolumn{1}{c}{\textbf{Level 1}} & \multicolumn{1}{c}{\textbf{Level 2}} & \multicolumn{1}{c}{\textbf{Level 3}} & \textbf{Total} 
         & \multicolumn{1}{c}{\textbf{Level 1}} & \multicolumn{1}{c}{\textbf{Level 2}} & \multicolumn{1}{c}{\textbf{Level 3}} & \textbf{Total} \\ 
             \midrule
VGPN~\cite{hu2018vgpn}    & \multicolumn{1}{c}{0.17 (\, 5/30)} & \multicolumn{1}{c}{0.13 (\, 4/30)} & \multicolumn{1}{c}{0.03 (\, 1/30)} & 0.11 (10/90)
         & \multicolumn{1}{c}{- (-)} & \multicolumn{1}{c}{- (-)} & \multicolumn{1}{c}{- (-)} & - (-) \\ 
ECRAP~\cite{oyama2024ecrap}    & \multicolumn{1}{c}{0.27 (\, 8/30)} & \multicolumn{1}{c}{0.27 (\, 8/30)} & \multicolumn{1}{c}{- (-)} & 0.27 (16/60)
         & \multicolumn{1}{c}{0.57 (17/30)} & \multicolumn{1}{c}{0.57 (17/30)} & \multicolumn{1}{c}{- (-)} & 0.57 (34/60) \\ 
MIEL (ours) & \multicolumn{1}{c}{\textbf{0.63 (19/30)}} & \multicolumn{1}{c}{\textbf{0.60 (18/30)}} & \multicolumn{1}{c}{\textbf{0.37 (11/30)}} & \textbf{0.53 (48/90)}
                & \multicolumn{1}{c}{\textbf{0.87 (26/30)}} & \multicolumn{1}{c}{\textbf{0.87 (26/30)}} & \multicolumn{1}{c}{\textbf{0.63 (19/30)}} & \textbf{0.79 (71/90)} \\ \hline
Human (topline)    & \multicolumn{1}{c}{1.00 (30/30)} & \multicolumn{1}{c}{1.00 (30/30)} & \multicolumn{1}{c}{0.93 (28/30)} & 0.98 (88/90)
         & \multicolumn{1}{c}{1.00 (30/30)} & \multicolumn{1}{c}{1.00 (30/30)} & \multicolumn{1}{c}{1.00 (30/30)} & 1.00 (90/90) \\ 
Human (w/o Q\&A) & \multicolumn{1}{c}{0.97 (29/30)} & \multicolumn{1}{c}{0.93 (28/30)} & \multicolumn{1}{c}{0.67 (20/30)} & 0.86 (77/90)
                 & \multicolumn{1}{c}{1.00 (30/30)} & \multicolumn{1}{c}{1.00 (30/30)} & \multicolumn{1}{c}{0.97 (29/30)} & 0.99 (89/90) \\ 
    \bottomrule
\end{tabular}
\end{table*}

\subsection{Evaluation Metrics}
\label{subsec:evaluation metric}
    The success rate (SR) was used as the evaluation metric.
    SR is calculated with this equation $\text{SR} = \frac{1}{N}\sum_{i=1}^N S_i$.
    $N$ is the number of attempts, $S_i$ is set to 1 if the $i$-th attempt succeeds and 0 if it fails.

    In this experiment, two types of SR were used: Top1 and Top5.
    For SR (Top1), success is defined when the target object is ranked first in the exophora resolution results, and the SR is then calculated.
    SR (Top5) is considered successful if the target object appears among the top five objects in the exophora resolution results, and the SR is likewise calculated.

\subsection{Results}
\label{subsec:exp_result_ER+FM,SSL}
    The results of this experiment are shown in Tables~\ref{tbl:result_each_method_combined_seen} and \ref{tbl:result_each_method_combined_unseen}.
    Table~\ref{tbl:result_each_method_combined_seen} shows the SR when the user is visible from the robot's initial position.
    Table~\ref{tbl:result_each_method_combined_unseen} shows the SR when the user is not visible.
    Table~\ref{tbl:result_ablation_study_combined} reports the results of an ablation study in which the SSL and interactive questioning modules were excluded from the MIEL (ours).
    ECRAP cannot predict the target object when both the target object's class information and the user's skeletal information are absent from the query.
    Therefore, the SR value for ECRAP at level~3 in Table~\ref{tbl:result_each_method_combined_unseen} is not reported.
    Since SSL succeeded in all trials during the experiment, the SR of the MIEL remains consistent across both Tables~\ref{tbl:result_each_method_combined_seen} and~\ref{tbl:result_each_method_combined_unseen}.

    Firstly, we compared the methods under the condition that the user is visible from the robot's initial position.
    When comparing the MIEL with ECRAP, the SR of the MIEL exceeds that of ECRAP by a factor of 1.3.
    For level~1 and level~2 queries that include object class information, there is little difference in SR between the two methods.
    However, for level~3 queries, which do not contain object class information, the MIEL achieves more than twice the SR of ECRAP.
    This suggests that the MIEL can compensate for the lack of information by interacting with the user, allowing it to perform a certain degree of exophora resolution even for level~3 queries that contain minimal information.
    The SR of Human (topline) was 0.98, while that of Human (w/o Q\&A) was 0.86, both demonstrating a very high SR.
    In comparison, the SR of the MIEL was 30–60\% lower than that of Human (topline) and Human (w/o Q\&A).
    Two factors are causing the large difference.
    Firstly, it was challenging to generate appropriate questions to narrow down the target object based on the given specific instructions.
    For example, when MIEL's exophora resolution failed to narrow down the object in response to the instruction ``Bring me the target object,'' GPT-4o repeatedly asked the user, ``What is the object used for?'' to gather additional information.
    Secondly, while MIEL estimates the target object based on the user's pointing, human participants may have also utilized gaze information in addition to pointing.
    This additional nonverbal cue could have contributed to the higher SR observed in human performance.

    Next, we compare the methods under the condition where the user is not visible from the robot's initial position.
    Under this condition, the MIEL outperforms ECRAP by a factor of three in terms of SR.
    Despite the user's invisibility, the MIEL achieves the same level of SR as when the user is visible by leveraging SSL to obtain the user's skeletal data.
    In contrast, when the user is not visible, ECRAP relies solely on the object class data in the query to predict the target object, leading to a significant drop in SR.
    Therefore, orienting the robot toward the user using SSL and compensating for the missing skeletal information is considered effective in maintaining stable SR in exophora resolution.

\begin{table*}[bt]
\centering
\caption{
The results of the ablation study for exophora resolution. L1, L2, and L3 indicate levels.
}
\label{tbl:result_ablation_study_combined}
\begin{tabular}{cc cccc cccc cccc cccc}
    \toprule
\multicolumn{2}{c}{\textbf{Modules}} & \multicolumn{8}{c}{\textbf{User Visible}} & \multicolumn{8}{c}{\textbf{User Not Visible}} \\
\multicolumn{2}{c}{} 
& \multicolumn{4}{c}{\textbf{Top 1}} & \multicolumn{4}{c}{\textbf{Top 5}} 
& \multicolumn{4}{c}{\textbf{Top 1}} & \multicolumn{4}{c}{\textbf{Top 5}} \\
\cmidrule(lr){3-6} \cmidrule(lr){7-10} \cmidrule(lr){11-14} \cmidrule(lr){15-18}
\textbf{SSL} & \textbf{Q\&A} 
& \textbf{L1} & \textbf{L2} & \textbf{L3} & \textbf{Total} 
& \textbf{L1} & \textbf{L2} & \textbf{L3} & \textbf{Total} 
& \textbf{L1} & \textbf{L2} & \textbf{L3} & \textbf{Total} 
& \textbf{L1} & \textbf{L2} & \textbf{L3} & \textbf{Total} \\
    \midrule
\checkmark &              
  & 0.53 & 0.40 & 0.13 & 0.36
  & 0.83 & 0.83 & 0.47 & 0.71
  & 0.53 & 0.40 & 0.13 & 0.36
  & 0.83 & 0.83 & 0.47 & 0.71 \\

        & \checkmark   
  & {\textbf{0.63}} & {\textbf{0.60}} & {\textbf{0.37}} & {\textbf{0.53}}
  & {\textbf{0.87}} & {\textbf{0.87}} & {\textbf{0.63}} & {\textbf{0.79}}
  & 0.40 & 0.37 & 0.20 & 0.32
  & 0.60 & 0.63 & 0.40 & 0.54 \\

\checkmark & \checkmark   
  & {\textbf{0.63}} & {\textbf{0.60}} & {\textbf{0.37}} & {\textbf{0.53}}
  & {\textbf{0.87}} & {\textbf{0.87}} & {\textbf{0.63}} & {\textbf{0.79}}
  & {\textbf{0.63}} & {\textbf{0.60}} & {\textbf{0.37}} & {\textbf{0.53}}
  & {\textbf{0.87}} & {\textbf{0.87}} & {\textbf{0.63}} & {\textbf{0.79}} \\
    \bottomrule
\end{tabular}
\end{table*}

    Finally, the results of the ablation study are used to examine the extent to which SSL and robot asking the user (Q\&A) contribute to exophora resolution in the MIEL.
    In the absence of Q\&A, there was only a slight difference in SR for level~1 queries.
    However, as the query level increased to level~2 and level~3, the difference in SR became more pronounced.
    For level 3 queries with no object class data or object feature words in the query, the SR with Q\&A was more than twice that of the query without Q\&A.
    These findings suggest that the less information a query contains about the target object, the more effective it is to supplement that information through Q\&A.
    Furthermore, the SR with SSL was approximately 1.7 times higher than that without SSL.
    These results support the effectiveness of the MIEL in obtaining skeletal data by reorienting toward the user based on SSL when such information is not initially available.
    

\subsection{Limitations}
\label{subsec:limitation}
    In this section, we outline several limitations of MIEL. 
    Firstly, our method requires interactive questioning to clarify ambiguities, increasing user burden and task completion time.
    Ideally, accurate exophora resolution would occur without user involvement.
    %
    Secondly, the SR of exophora resolution relies heavily on the quality of the semantic map.
    Objects absent from the map or low-resolution images may cause identification failures.
    %
    Thirdly, MIEL has difficulty adapting to changes in object movement after semantic mapping, which leads to failures in exophora resolution.
    %
    Lastly, the SR of SSL may significantly decrease due to poor visibility or noise.
    If SSL fails, it could lead to a substantial reduction in the SR of exophora resolution.

\section{Conclusion}
\label{sec:conclusion}
    In this paper, we proposed a new method for exophora resolution, called MIEL, which leverages SSL and interactive questioning by LLMs to achieve robust performance even in the presence of missing observational information.
    In the experiments, MIEL was compared against a baseline method that lacked both the interactive questioning module and SSL, as well as a topline approach involving human prediction.
    MIEL achieved approximately 1.3 times the SR of the method of exophora resolution without SSL and interactive questioning when the user was visible from the robot's initial position.
    Moreover, it was about 2.0 times the SR when the user was not visible.
    

    In future works, we will address the four challenges listed in Section~\ref{subsec:limitation}.
    The first challenge is increased user burden due to interactive questioning, which we address by refining the question generation method to maintain the number of questions while achieving the desired results.
    The second challenge is the degradation of semantic map quality due to low-resolution object images, which we tackle by applying super-resolution.
    The third challenge is handling dynamic object arrangements, which we address by integrating a real-time semantic map update method.
    The fourth challenge is exophora resolution in noisy environments, which we solve by using sound source separation to distinguish a specific person's voice from background noise.



\bibliographystyle{templates/IEEEtran}
\bibliography{root}

\end{document}